\documentclass{article}
\usepackage{ijcai21}

\usepackage{times}

\usepackage{soul}
\usepackage{url}
\usepackage[hidelinks]{hyperref}
\usepackage[utf8]{inputenc}
\usepackage[small]{caption}
\usepackage{graphicx}
\usepackage{amsmath}
\usepackage{booktabs}
\usepackage{multirow}
\title{Federated Graph Learning - A Position Paper}

\author{
Huanding Zhang\footnote{Equal contributions}$^1$\and
Tao Shen\footnotemark[1]$^3$\and
Fei Wu$^3$\and
Mingyang Yin$^4$\and
Hongxia Yang$^4$\And
Chao Wu\footnote{Contact Author}$^{2}$\\
\affiliations
$^1$School of Software Technology, Zhejiang University, Hangzhou, China\\
$^2$School of Public Affairs, Zhejiang University, Hangzhou, China\\
$^3$Department of Computer Science, Zhejiang University, Hangzhou, China\\
$^4$DAMO Academy, Alibaba Group, Hangzhou, China\\
\emails
\{zhanghuanding, tao.shen, wufei, chao.wu\}@zju.edu.cn,
\{hengyang.ymy, yang.yhx\}@alibaba-inc.com
}

\begin{document}

\maketitle

\begin{abstract}
    Graph neural networks (GNN) have been successful in many fields, and derived various researches and applications in real industries. However, in some privacy sensitive scenarios (like finance, healthcare), training a GNN model centrally faces challenges due to the distributed data silos. Federated learning (FL) is a an emerging technique that can collaboratively train a shared model while keeping the data decentralized, which is a rational solution for distributed GNN training. We term it as federated graph learning (FGL). Although FGL has received increasing attention recently, the definition and challenges of FGL is still up in the air. In this position paper, we present a categorization to clarify it. Considering how graph data are distributed among clients, we propose four types of FGL: inter-graph FL, intra-graph FL and graph-structured FL, where intra-graph is further divided into horizontal and vertical FGL. For each type of FGL, we make a detailed discussion about the formulation and applications, and propose some potential challenges.
\end{abstract}

\section{Motivation}
    Graph neural networks (GNN) have demonstrated remarkable performance in modeling graph data, and derived various researches and applications in real industries like finance \cite{liu2018heterogeneous} \cite{liu2019geniepath} \cite{wang2019semi}, traffic \cite{yu2017spatio}, recommender systems \cite{ying2018graph}, chemistry \cite{wang2020gognn}, etc. However, GNN still faces many problems and one of them is the data silos. Because of the privacy concern or commercial competition, data exist in a isolated manner, giving rise to challenges on centrally training GNN. For example, banks may leverage GNN as anti-fraud models, but they only have transactions data of locally registered users (subgraph), thus the model is not effective for other users. Also, pharmaceutical companies usually utilize GNN for drug discovery and synthesis, while the data are quite limited and confidential in independent research institution of companies. Whereas GNN has been successful in many fields, isolated data restrict its further development.

    Federated learning (FL) is a machine learning setting where clients can collaboratively train a shared model under the orchestration of central server, while keeping the data decentralized. Unlike traditional centralized machine learning techniques, data are fixed locally, rather than being gathered in central server, who exists many of the systemic privacy risks and costs \cite{kairouz2019advances}. Back to aforementioned examples, with federated learning, banks or pharmaceutical companies can collaboratively train a shared GNN model, utilize isolated data while keeping them safe and local. Hence, FL is a promising solution for training GNN over isolated graph data, and in this paper we term it as federated graph learning (FGL).

    As far as we know, FGL has received increasing attention recently. \cite{zheng2020asfgnn} devises a novel FL framework for GNN that supports automatically hyper-parameters optimization. \cite{wang2020graphfl} proposes a FL framework for semi-supervised node classification based on meta learning. \cite{jiang2020federated} presents a method to learn dynamic representation of objects from multi-user graph sequences. \cite{wu2021fedgnn} designs a federated GNN framework for privacy preserving recommendation. \cite{scardapane2020distributed} presents a distributed training method for GNN, but it preserves the edges among subgraphs.

    However, the definition and challenges of FGL is still up in the air. Although \cite{he2021fedgraphnn} proposes a rather comprehensive benchmark for FGL, it is not detailed enough about categorization. In this position paper, we present a categorization to clarify it. Considering how graph data are distributed among clients, we propose four types of FGL: inter-graph FL, intra-graph FL and graph-structured FL, where intra-graph is further divided into horizontal and vertical FGL, referring to the categorization of FL \cite{yang2019federated}. For each type of FGL, we discuss the formulation, applications and challenges. The rest of this paper is organized as follows: In Section \ref{categorization}, we detail four types of FGL. In Section \ref{challenges}, we analysis potential challenges and possible solution for each type of FGL.

\section{A categorization of federated graph learning}
\label{categorization}
    We introduce four types of FGL from the perspective of how graph data are distributed in FL. They are summarized in table \ref{Type_of_FGL_table}, details will be discussed as follows. Without loss of generality, we follow the settings of Graph Convolutional Networks (GCN) \cite{kipf2016semi} and Federated Averaging (FedAvg) \cite{mcmahan2017communication} for convenience.

    A typical FL framework consists of a server and $K$ clients. The $k^{th}$ client has its own dataset $ D_k$ with size of $|D_k| = N_k$, and $N = \sum_{k=1}^{K} N_k$. Graph convolution variants \cite{velivckovic2017graph} \cite{hamilton2017inductive} can be generally formulated as the MPNN framework:
    
    \begin{equation}
		x^{l}_{i} = \gamma^{l}(x^{l-1}_{i}, Aggr^{l}_{ j \in \mathcal{ N }_{i}} \phi^{l} (x^{l - 1}_{i}, x^{l - 1}_{j}, a_{ij} ) ), \label{GCN_layer}
	  \end{equation}

    where the graph is $\mathcal{G}=(\mathcal{V}, \mathcal{E})$, $x^{l}_{i}$ is the $i^{th}$ node feature in $l^{th}$ layer, $a_{ij}$ is the edge feature between node $i$ and node $j$, $\mathcal{N}$ denotes the neighbor set of node $i$, $Aggr$ denotes differentiable aggregation function($sum$, $mean$, $max$, etc.), $\gamma$ and $\phi$ denote differentiable function (e.g. MLP). For simplicity, a GCN model composed by \ref{GCN_layer} can be denoted as $ H(X, A, W)$, where $X$ is feature matrix, $A$ is adjacency matrix, and  $W$ denotes parameters.

\begin{figure}[htp]
    \centering
    \includegraphics[width=0.48\textwidth]{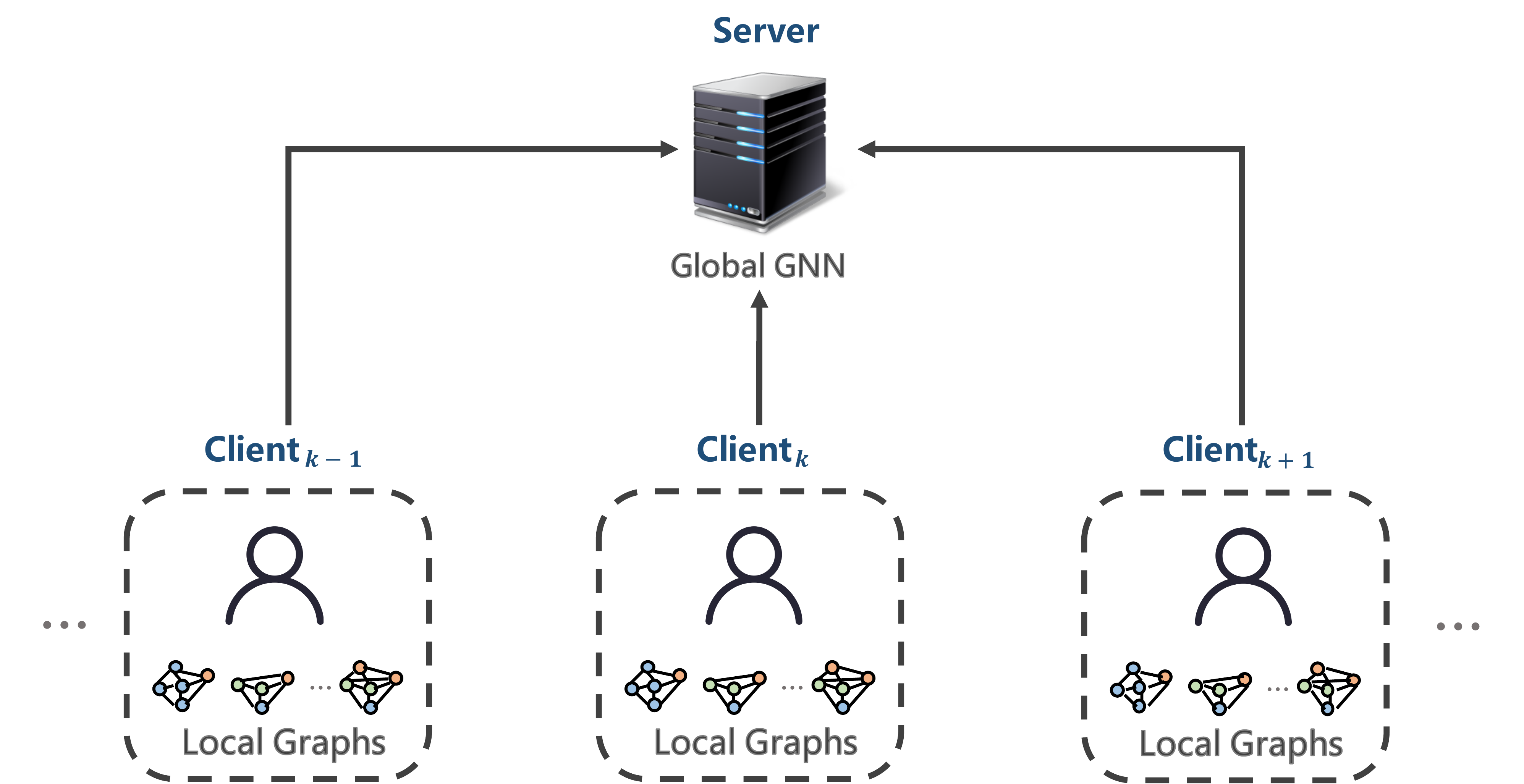}
    \caption{Framework of inter-graph FL: the sample granularity is graph and global GNN model performs graph-level task.}
    \label{inter_graph_FL_plot}
\end{figure}

    \subsection{Inter-graph federated learning}
        This type of FGL is the most natural derivation of FL, where each sample of clients is of graph data, and global model performs graph-level task (shown as figure \ref{inter_graph_FL_plot}). The most typical application of inter-graph FL is in the biochemical industry where researchers use GNN to study the graph structure of molecules. A molecule can be represented as a graph where atoms are nodes and chemical bonds are edges. In the study of drug properties, every pharmaceutical company holds a confidential dataset $D_k$ which contains molecule structure $\{\mathcal{G}_i\}$ and corresponding properties $\{y_i\}$. In the past, commercial competition hindered their cooperation, but it becomes possible with the framework of inter-graph FL. 
        Under this setting, $D_k=\{ (\mathcal{G}^{(k)}_i, y^{(k)}_i) \}$, global model is
	    
        \begin{equation}
		    \hat{y}_{i}^{(k)} = H(X_{i}^{(k)}, A_{i}^{(k)}, W), 
	      \end{equation}

        where $X_{i}^{(k)}$ and $A_{i}^{(k)}$ denote feature and adjacency matrix of $i^{th}$ graph in $k^{th}$ client's dataset, $\hat{y}$ is output.
        
        Applying FedAvg, the objective function is

        \begin{equation}
        \begin{aligned}
            &\mathop{min} \limits_{W} \frac{N_{k}}{N} \sum_{k=1}^{K} f_{k}(W), \\ 
            f_{k}(W) =  \frac{1}{N_{k}} & \sum_{i=1}^{N_{k}} \mathcal{L} (H(X_{i}^{(k)}, A_{i}^{(k)}, W), y_{i}^{(k)}),
        \end{aligned}
        \end{equation}

        where $f_{k}(W)$ denotes local objective function and $\mathcal{L}$ is global loss. Pharmaceutical companies thus can collaboratively train a shared model without providing confidential data.

\begin{figure}[htp]
    \centering
    \includegraphics[width=0.45\textwidth]{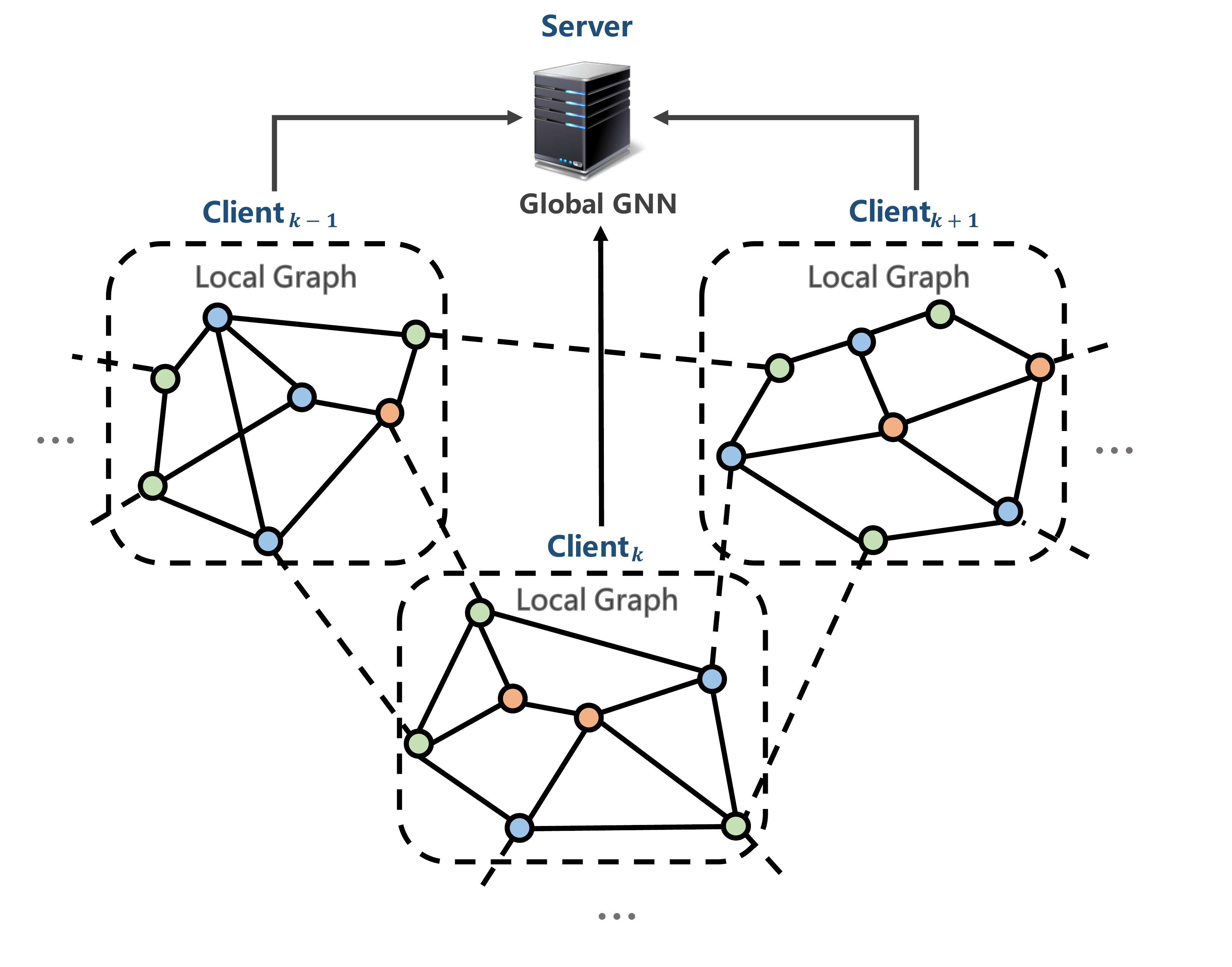}
    \caption{Framework of horizontal intra-graph FL:  subgraphs held in clients are horizontally distributed, edges represented as dashed line are the connections that should have been there but are missing.}
    \label{H_intra_graph_FL_plot}
\end{figure}

    \subsection{Intra-graph federated learning}
        Another type of FGL is the intra-graph federated learning, where each client own a part of latent entire graph. Referring to \cite{yang2019federated}, intra-graph federated learning can also be divided into horizontal and vertical FGL, corresponding to users and features who is partitioned.

        \subsubsection{Horizontal intra-graph FL}
            In this situation, the subgraphs held in each client appear to be horizontally partitioned from the latent entire graph(shown as in figure \ref{H_intra_graph_FL_plot}, connections among them are lost because of data isolated storage, strictly speaking, there can be overlap), that is, $A \Rightarrow \{ A^{(k)}\}$. Horizontally distributed subgraphs have the same properties, clients share the same feature and label space but different node ID space. Under this setting, $D_k=(\mathcal{G}^{(k)}, Y^{(k)})$, $N_k$ denotes the number of nodes in $\mathcal{G}^{(k)}$. Global GNN model performs node or link-level task,
            \begin{equation}
                \hat{Y}^{(k)} = H(X^{(k)}, A^{(k)}, W).
            \end{equation}
            The objective function becomes

	        \begin{equation}
            \begin{aligned}
		        &\mathop{min} \limits_{W} \frac{N_{k}}{N} \sum_{k=1}^{K} f_{k}(W), \\
                 f_{k}(W) = &\mathcal{L} (H(X^{(k)}, A^{(k)}, W), Y^{(k)}).
            \end{aligned}
            \end{equation}

            Subgraph horizontal distribution is very common in real world. For example, in online social app, each user has a local social network $\mathcal{G}^{(k)}$ and $\{ \mathcal{G}^{(k)}\}$ constitute the latent entire human social network $ \mathcal{G} $.  The developers are able to devise friend recommendation algorithm based on horizontal intra-graph FL to avoid violating users' social privacy.

\begin{figure}[htp]
    \centering
    \includegraphics[width=0.42\textwidth]{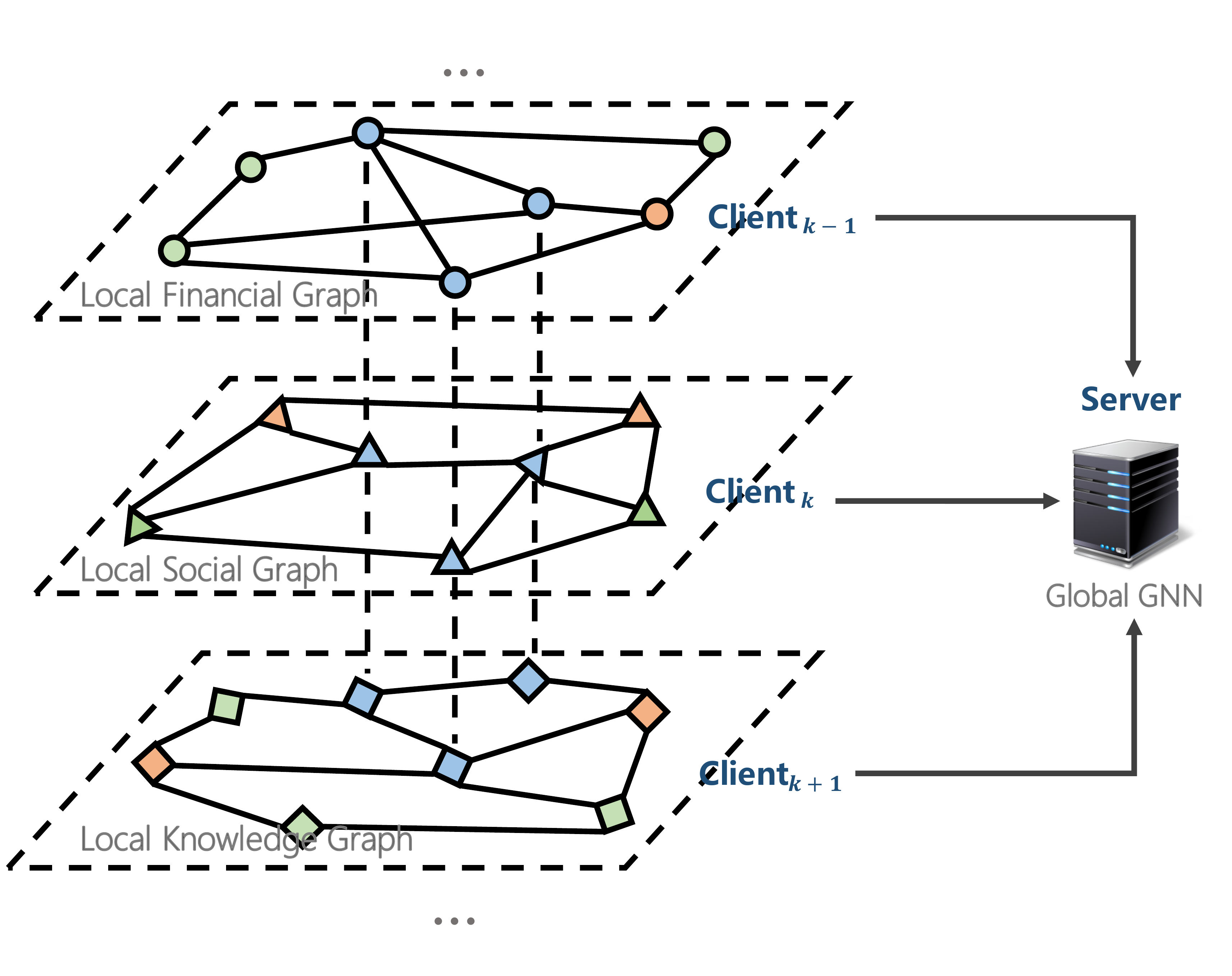}
    \caption{Framework of vertical intra-graph FL:  subgraphs held in clients are vertically distributed, and they are parallel and heavily overlap with each other, vertical dashed lines indicate the corresponding nodes have same ID.}
    \label{V_intra_graph_FL_plot}
\end{figure}

        \subsubsection{Vertical intra-graph FL}
            Subgraph vertical distribution means that they are parallel and heavily overlap with each other(shown as figure \ref{V_intra_graph_FL_plot}, graphs of financial, social and knowledge are vertically distributed). It's like the latent entire graph is vertically partitioned, that is, $A \Rightarrow \{A^{(k)}\}$, $X \Rightarrow \{X^{(k)}\}$, $Y \Rightarrow \{Y^{(k)}\}$. Under this setting, clients share the same node ID space but different feature and label space, $D_k=(\mathcal{G}^{(k)}, Y^{(k)})$, $V^{'}$ is set of common nodes, $N_k$ is size of $V'$. Global model is not unique(it depends on how many clients have labels), which indicates vertical intra-graph FL supports multi-task learning. The main purpose of vertical intra-graph FL is to learn more comprehensive GNN by combining $\{X^{(k)}_v | v \in V^{'}\}$  and sharing $\{Y^{(k)}_v | v \in V^{'}\}$ in a privacy preserved and communication efficient manner. Without considering the method of entity matching and data sharing, the objective function can be expressed as

            \begin{equation}
            \begin{aligned}
                &\mathop{min} \limits_{W}  f_{k}(W), \\
                f_{k}(W) =  \mathcal{L} (H(Aggr&_{k=1}^{K}(X_{V'}^{(k)}), A_{V'}^{(k)}, W^{(k)}), Y_{V'}^{(k)}).
            \end{aligned}
            \end{equation}
    
            Vertical intra-graph FL can be applied in the cooperation among organizations. For example, in detection of money laundering, criminals are tends to devise sophisticated strategies that span across different organizations. Due to privacy concern, banks need to hand over list of suspects to a trustworthy national institution and rely on them to do analysis. This procedure is inefficient. With the framework of vertical intra-graph FL, banks are able to collaboratively monitor money laundering activities in real-time while keeping their users' data protected. Some researchers have studied this type of FGL, \cite{suzumura2019towards} and \cite{chen2020fede} respectively devise a vertical intra-graph FL framework for financial fraud detection and knowledge graph embedding.
    
\begin{figure}[htp]
    \centering
    \includegraphics[width=0.42\textwidth]{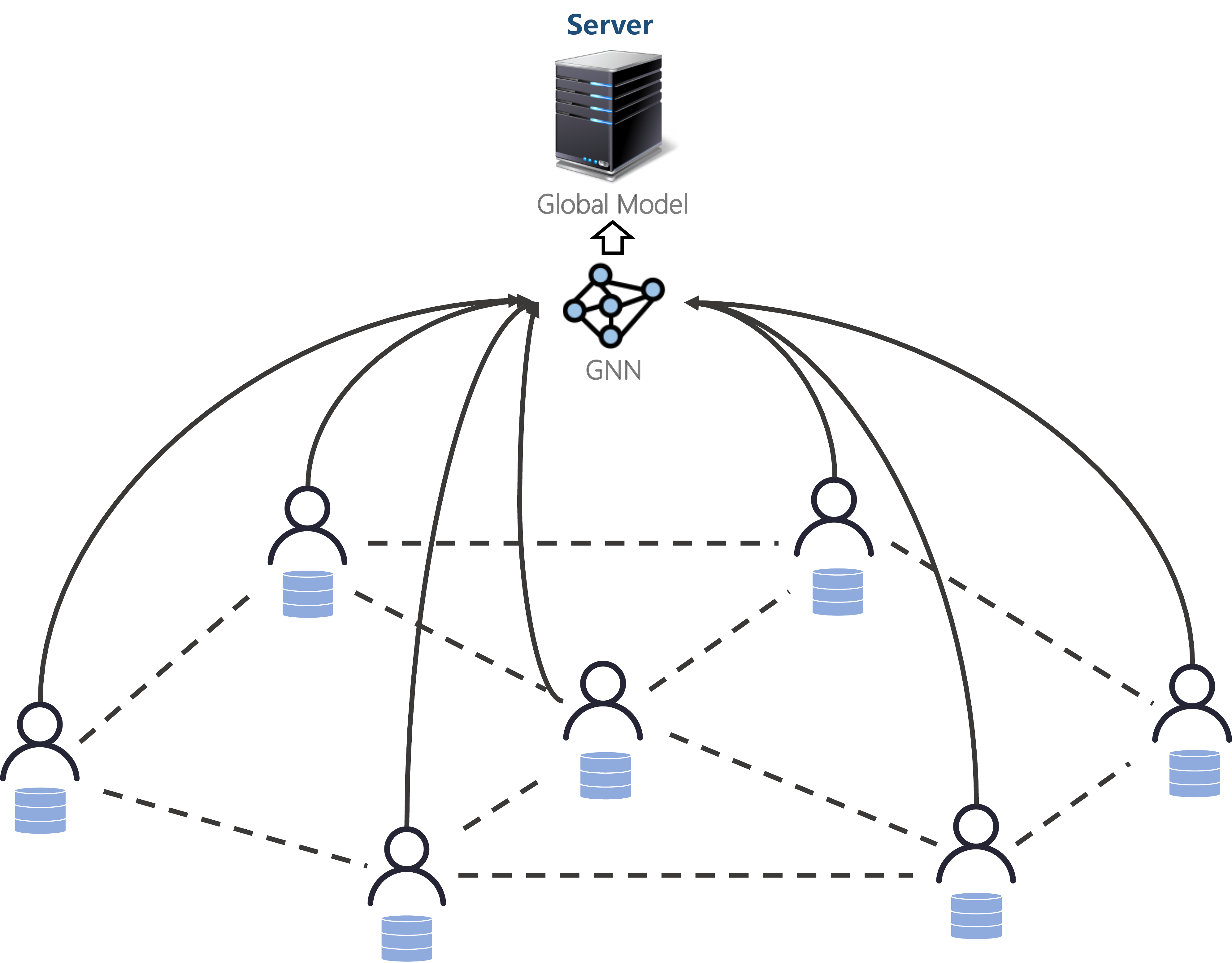}
    \caption{Framework of graph-structured FL: graphs exist as relationships among clients, GNN is used to extract inherent information from the topology of clients.}
    \label{graph_structed_FL_plot}
\end{figure}

    \subsection{Graph-structured federated learning}
        In addition to being data, graphs can also exit as relationships among clients (e.g. geography or social networks shown as figure \ref{graph_structed_FL_plot}), that is, $client^{(k)} \Rightarrow v_k$. Graph-structured federated learning refers to the situation where server uses GNN to aggregate local models based on clients topology.

        \begin{equation}
            f(w)^{(t)} = Aggr_{k=1}^{K}(f(w)^{(k, t-1)}, A, W).
        \end{equation}

        Under this setting, clients hold either Euclidean or graph data, global model performs any kind of task, objective function is the same as in FL. Graph-structured FL can be considered as a special federated optimization method since GNN is applied to extract inherent information among clients to improve FL. The typical application is in federated traffic flow prediction where monitoring devices are distributed in different geographic positions, GNN are used to capture spatial dependencies among devices \cite{meng2021crossnode}.
    
\renewcommand\arraystretch{1.5}
\begin{table*}[tbp]
    \centering
    \begin{tabular}{cccclc}
        \toprule
        \multicolumn{2}{c}{Type}           & Federalization & \multicolumn{2}{c}{Data form in clients}               & Global model task  \\
        
        \hline
        \multicolumn{2}{c}{Inter-graph Federated Learning}           & $D_k=\{\mathcal{G}_{i}\}$               & \multicolumn{2}{c}{ Graphs}                    & Graph-level        \\
        
        \hline
        \multirow{2}{*}{Intra-graph federated learning} & Horizontal & $A \Rightarrow \{ A^{(k)}\}$               & \multicolumn{2}{c}{Horizontally distributed subgraphs} & Node or link-level \\ 
        
        \cmidrule(r){2-6}
        & Vertical   &   \parbox{2cm}{$A \Rightarrow \{ A^{(k)}\}$ $X \Rightarrow \{ X^{(k)} \}$ $Y \Rightarrow \{ Y^{(k)}\}$}             & \multicolumn{2}{c}{Vertically distributed subgraphs}   & Node or link-level \\
        
        \hline
        \multicolumn{2}{c}{Graph-structured federated learning}           & $client^{(k)} \Rightarrow v_{k}$              & \multicolumn{2}{c}{Arbitrary}                          & Arbitrary \\        
        \bottomrule
    \end{tabular}
    \caption{Four types of FGL: each type corresponds to a different way of federalization of graph.}
    \label{Type_of_FGL_table}
\end{table*}

\section{Challenges}
\label{challenges}

    Though researchers have already proposed several FGL frameworks, there are still many problems. Most of them are left from FL and become more complicated in graph domain, such as Non-IID data, communication efficiency and robustness \cite{kairouz2019advances}. In this section, we discuss the main challenges and possible solutions for each type of FGL.

    \subsection{Non-IID graph structure}
        Non-IID problem is inevitable no matter in which type of FGL. Same as in FL, it can both impact convergence speed and accuracy. Researchers have attempted to devise some methods to alleviate its influence \cite{zheng2020asfgnn} \cite{wang2020graphfl}, as far as we know, there is no work solving it completely. In addition to feature and label, graph data have edge (structure) information, which indicates Non-IID of graph structure might influence the learning process as well. Properties of graph structure include degree distribution, average path length, average clustering coefficient, etc. Studying Non-IID of these properties might be an important aspect of solving Non-IID problem in graph domain. No work has paid attention to studying Non-IID of graph structure yet, and it's worth digging.

    \subsection{Isolated graph in horizontal intra-graph FL}
        Representation learning on graph models relies on walking or message passing through multi-order neighbors. However, the latent entire graph is isolated by different data holders in horizontal intra-graph FL, the diameter of local subgraph is nearly small. It will impact the accuracy of GCN since the local graph cannot provide information from high-order neighbors. Consider an extreme case where the local subgraph only contains one node, GCN degenerates to MLP. Therefore, discovering latent edges among local subgraphs of clients is a crucial challenge in horizontal intra-graph FL, there are some researches who have mentioned it, \cite{wu2021fedgnn} proposes a method based on homomorphic encryption to expand local subgraph, more ideas are needed on this issue.

    \subsection{Entities matching and secure data sharing in vertical intra-graph FL}
        Entities matching and secure data sharing are key problems for both vertical FL and vertical intra-graph FL. \cite{hardy2017private} achieves learning a federated logistic regression model between two participants based on additively homomorphic encryption, and \cite{feng2020multi} generalizes it to multi-participants and multi-class classification. Vertical intra-graph FL is at least as complicated as VFL, and the main difficulties also lie in ensuring precision, privacy preserving and communication efficient at the same time. There is no vertical intra-graph FL framework achieving these requires. \cite{chen2020fede} proposes a federated framework to do knowledge graph embedding by a matching table held in server, which violates privacy preserving to some extent.

    \subsection{Dataset of intra-graph FL}
        The richness of image and corpus dataset is a necessary condition for rapid development of computer vision and natural language processing. However, there is no suitable graph dataset for intra-graph FL. For Euclidean data, we can easily simulate data distribution in experiments. However, simulation becomes difficult when it comes to graph data due to the additional structure information. For example, in horizontal setting, we have to split a graph into multiple subgraphs but the removed edges and subgraph distribution are not in line with reality. It can be hard in vertical setting as well. Although features can be split into several partitions, whether all partitions have the same structure needs to be considered. It is usually more complicated in real scenes. Thus, the lack of datasets limits the development of intra-graph FL.
    
    \subsection{Communication and memory consumption}
        Communication and memory consumption turns out to be a key bottleneck when applying federated algorithms in reality. For example, for federated recommender system, models transmitted between server and clients may be heavy, where user/item representation layers occupy most of model parameters and the size of representation parameters grows linearly with the ever-increasing scale of user/item. It brings unfavorable both communication and memory consumption. Model quantization, pruning, distillation are effective methods for model compression. \cite{tailor2020degree} studies model quantization method for GNN. \cite{yang2020distilling} proposes a distillation approach which transfers topology-aware knowledge from teacher GCN to student GCN. \cite{lian2020lightrec} devises an end-to-end framework for learning quantization of item representation in recommender system. Thus, compression technique for GNN  is also a potential way for FGL.


\begin{thebibliography}{}

    \bibitem[\protect\citeauthoryear{Chen \bgroup \em et al.\egroup
      }{2020}]{chen2020fede}
    Mingyang Chen, Wen Zhang, Zonggang Yuan, Yantao Jia, and Huajun Chen.
    \newblock Fede: Embedding knowledge graphs in federated setting.
    \newblock {\em arXiv preprint arXiv:2010.12882}, 2020.
    
    \bibitem[\protect\citeauthoryear{Feng and Yu}{2020}]{feng2020multi}
    Siwei Feng and Han Yu.
    \newblock Multi-participant multi-class vertical federated learning.
    \newblock {\em arXiv preprint arXiv:2001.11154}, 2020.
    
    \bibitem[\protect\citeauthoryear{Hamilton \bgroup \em et al.\egroup
      }{2017}]{hamilton2017inductive}
    William~L Hamilton, Rex Ying, and Jure Leskovec.
    \newblock Inductive representation learning on large graphs.
    \newblock {\em arXiv preprint arXiv:1706.02216}, 2017.
    
    \bibitem[\protect\citeauthoryear{Hardy \bgroup \em et al.\egroup
      }{2017}]{hardy2017private}
    Stephen Hardy, Wilko Henecka, Hamish Ivey-Law, Richard Nock, Giorgio Patrini,
      Guillaume Smith, and Brian Thorne.
    \newblock Private federated learning on vertically partitioned data via entity
      resolution and additively homomorphic encryption.
    \newblock {\em arXiv preprint arXiv:1711.10677}, 2017.
    
    \bibitem[\protect\citeauthoryear{He \bgroup \em et al.\egroup
      }{2021}]{he2021fedgraphnn}
    Chaoyang He, Keshav Balasubramanian, Emir Ceyani, Yu~Rong, Peilin Zhao, Junzhou
      Huang, Murali Annavaram, and Salman Avestimehr.
    \newblock Fedgraphnn: A federated learning system and benchmark for graph
      neural networks.
    \newblock {\em arXiv preprint arXiv:2104.07145}, 2021.
    
    \bibitem[\protect\citeauthoryear{Jiang \bgroup \em et al.\egroup
      }{2020}]{jiang2020federated}
    Meng Jiang, Taeho Jung, Ryan Karl, and Tong Zhao.
    \newblock Federated dynamic gnn with secure aggregation.
    \newblock {\em arXiv preprint arXiv:2009.07351}, 2020.
    
    \bibitem[\protect\citeauthoryear{Kairouz \bgroup \em et al.\egroup
      }{2019}]{kairouz2019advances}
    Peter Kairouz, H~Brendan McMahan, Brendan Avent, Aur{\'e}lien Bellet, Mehdi
      Bennis, Arjun~Nitin Bhagoji, Keith Bonawitz, Zachary Charles, Graham Cormode,
      Rachel Cummings, et~al.
    \newblock Advances and open problems in federated learning.
    \newblock {\em arXiv preprint arXiv:1912.04977}, 2019.
    
    \bibitem[\protect\citeauthoryear{Kipf and Welling}{2016}]{kipf2016semi}
    Thomas~N Kipf and Max Welling.
    \newblock Semi-supervised classification with graph convolutional networks.
    \newblock {\em arXiv preprint arXiv:1609.02907}, 2016.
    
    \bibitem[\protect\citeauthoryear{Lian \bgroup \em et al.\egroup
      }{2020}]{lian2020lightrec}
    Defu Lian, Haoyu Wang, Zheng Liu, Jianxun Lian, Enhong Chen, and Xing Xie.
    \newblock Lightrec: A memory and search-efficient recommender system.
    \newblock In {\em Proceedings of The Web Conference 2020}, pages 695--705,
      2020.
    
    \bibitem[\protect\citeauthoryear{Liu \bgroup \em et al.\egroup
      }{2018}]{liu2018heterogeneous}
    Ziqi Liu, Chaochao Chen, Xinxing Yang, Jun Zhou, Xiaolong Li, and Le~Song.
    \newblock Heterogeneous graph neural networks for malicious account detection.
    \newblock In {\em Proceedings of the 27th ACM International Conference on
      Information and Knowledge Management}, pages 2077--2085, 2018.
    
    \bibitem[\protect\citeauthoryear{Liu \bgroup \em et al.\egroup
      }{2019}]{liu2019geniepath}
    Ziqi Liu, Chaochao Chen, Longfei Li, Jun Zhou, Xiaolong Li, Le~Song, and Yuan
      Qi.
    \newblock Geniepath: Graph neural networks with adaptive receptive paths.
    \newblock In {\em Proceedings of the AAAI Conference on Artificial
      Intelligence}, volume~33, pages 4424--4431, 2019.
    
    \bibitem[\protect\citeauthoryear{McMahan \bgroup \em et al.\egroup
      }{2017}]{mcmahan2017communication}
    Brendan McMahan, Eider Moore, Daniel Ramage, Seth Hampson, and Blaise~Aguera
      y~Arcas.
    \newblock Communication-efficient learning of deep networks from decentralized
      data.
    \newblock In {\em Artificial Intelligence and Statistics}, pages 1273--1282.
      PMLR, 2017.
    
    \bibitem[\protect\citeauthoryear{Meng \bgroup \em et al.\egroup
      }{2021}]{meng2021crossnode}
    Chuizheng Meng, Sirisha Rambhatla, and Yan Liu.
    \newblock Cross-node federated graph neural network for spatio-temporal data
      modeling, 2021.
    
    \bibitem[\protect\citeauthoryear{Scardapane \bgroup \em et al.\egroup
      }{2020}]{scardapane2020distributed}
    Simone Scardapane, Indro Spinelli, and Paolo Di~Lorenzo.
    \newblock Distributed graph convolutional networks.
    \newblock {\em arXiv preprint arXiv:2007.06281}, 2020.
    
    \bibitem[\protect\citeauthoryear{Suzumura \bgroup \em et al.\egroup
      }{2019}]{suzumura2019towards}
    Toyotaro Suzumura, Yi~Zhou, Natahalie Baracaldo, Guangnan Ye, Keith Houck, Ryo
      Kawahara, Ali Anwar, Lucia~Larise Stavarache, Yuji Watanabe, Pablo Loyola,
      et~al.
    \newblock Towards federated graph learning for collaborative financial crimes
      detection.
    \newblock {\em arXiv preprint arXiv:1909.12946}, 2019.
    
    \bibitem[\protect\citeauthoryear{Tailor \bgroup \em et al.\egroup
      }{2020}]{tailor2020degree}
    Shyam~A Tailor, Javier Fernandez-Marques, and Nicholas~D Lane.
    \newblock Degree-quant: Quantization-aware training for graph neural networks.
    \newblock {\em arXiv preprint arXiv:2008.05000}, 2020.
    
    \bibitem[\protect\citeauthoryear{Veli{\v{c}}kovi{\'c} \bgroup \em et al.\egroup
      }{2017}]{velivckovic2017graph}
    Petar Veli{\v{c}}kovi{\'c}, Guillem Cucurull, Arantxa Casanova, Adriana Romero,
      Pietro Lio, and Yoshua Bengio.
    \newblock Graph attention networks.
    \newblock {\em arXiv preprint arXiv:1710.10903}, 2017.
    
    \bibitem[\protect\citeauthoryear{Wang \bgroup \em et al.\egroup
      }{2019}]{wang2019semi}
    Daixin Wang, Jianbin Lin, Peng Cui, Quanhui Jia, Zhen Wang, Yanming Fang, Quan
      Yu, Jun Zhou, Shuang Yang, and Yuan Qi.
    \newblock A semi-supervised graph attentive network for financial fraud
      detection.
    \newblock In {\em 2019 IEEE International Conference on Data Mining (ICDM)},
      pages 598--607. IEEE, 2019.
    
    \bibitem[\protect\citeauthoryear{Wang \bgroup \em et al.\egroup
      }{2020a}]{wang2020graphfl}
    Binghui Wang, Ang Li, Hai Li, and Yiran Chen.
    \newblock Graphfl: A federated learning framework for semi-supervised node
      classification on graphs.
    \newblock {\em arXiv preprint arXiv:2012.04187}, 2020.
    
    \bibitem[\protect\citeauthoryear{Wang \bgroup \em et al.\egroup
      }{2020b}]{wang2020gognn}
    Hanchen Wang, Defu Lian, Ying Zhang, Lu~Qin, and Xuemin Lin.
    \newblock Gognn: Graph of graphs neural network for predicting structured
      entity interactions.
    \newblock {\em arXiv preprint arXiv:2005.05537}, 2020.
    
    \bibitem[\protect\citeauthoryear{Wu \bgroup \em et al.\egroup
      }{2021}]{wu2021fedgnn}
    Chuhan Wu, Fangzhao Wu, Yang Cao, Yongfeng Huang, and Xing Xie.
    \newblock Fedgnn: Federated graph neural network for privacy-preserving
      recommendation.
    \newblock {\em arXiv preprint arXiv:2102.04925}, 2021.
    
    \bibitem[\protect\citeauthoryear{Yang \bgroup \em et al.\egroup
      }{2019}]{yang2019federated}
    Qiang Yang, Yang Liu, Tianjian Chen, and Yongxin Tong.
    \newblock Federated machine learning: Concept and applications.
    \newblock {\em ACM Transactions on Intelligent Systems and Technology (TIST)},
      10(2):1--19, 2019.
    
    \bibitem[\protect\citeauthoryear{Yang \bgroup \em et al.\egroup
      }{2020}]{yang2020distilling}
    Yiding Yang, Jiayan Qiu, Mingli Song, Dacheng Tao, and Xinchao Wang.
    \newblock Distilling knowledge from graph convolutional networks.
    \newblock In {\em Proceedings of the IEEE/CVF Conference on Computer Vision and
      Pattern Recognition}, pages 7074--7083, 2020.
    
    \bibitem[\protect\citeauthoryear{Ying \bgroup \em et al.\egroup
      }{2018}]{ying2018graph}
    Rex Ying, Ruining He, Kaifeng Chen, Pong Eksombatchai, William~L Hamilton, and
      Jure Leskovec.
    \newblock Graph convolutional neural networks for web-scale recommender
      systems.
    \newblock In {\em Proceedings of the 24th ACM SIGKDD International Conference
      on Knowledge Discovery \& Data Mining}, pages 974--983, 2018.
    
    \bibitem[\protect\citeauthoryear{Yu \bgroup \em et al.\egroup
      }{2017}]{yu2017spatio}
    Bing Yu, Haoteng Yin, and Zhanxing Zhu.
    \newblock Spatio-temporal graph convolutional networks: A deep learning
      framework for traffic forecasting.
    \newblock {\em arXiv preprint arXiv:1709.04875}, 2017.
    
    \bibitem[\protect\citeauthoryear{Zheng \bgroup \em et al.\egroup
      }{2020}]{zheng2020asfgnn}
    Longfei Zheng, Jun Zhou, Chaochao Chen, Bingzhe Wu, Li~Wang, and Benyu Zhang.
    \newblock Asfgnn: Automated separated-federated graph neural network.
    \newblock {\em arXiv preprint arXiv:2011.03248}, 2020.
    
    \end{thebibliography}

\end{document}